\documentclass[runningheads]{llncs}
\usepackage{graphicx}
\usepackage{amsmath,amssymb} 
\usepackage{color}
\usepackage{amsmath}
\usepackage{amsfonts}
\usepackage{lipsum}
\usepackage{float}
\usepackage{wrapfig, blindtext}
\usepackage[hidelinks]{hyperref}
\usepackage{siunitx}
\usepackage{graphicx}
\usepackage{amssymb}
\usepackage{subfig}
\usepackage{multirow}
\graphicspath{{images/}}
\usepackage{comment}
\usepackage{textcomp}
\usepackage{color}
\usepackage{rotating}
\usepackage{soul}
\setstcolor{red}
\usepackage[utf8x]{inputenc}
\DeclareRobustCommand*{\ora}{\overrightarrow}
\usepackage[table]{xcolor}
\usepackage{sidecap,wasysym,array,tabularx,caption}
\usepackage{booktabs}
\definecolor{mygray}{gray}{.9}
\begin{document}
\pagestyle{headings}
\mainmatter

\title{`Labelling the Gaps': A Weakly Supervised Automatic Eye Gaze Estimation} 

\author{Shreya Ghosh$^1$, Abhinav Dhall$^{1,2}$, Jarrod Knibbe$^3$ and Munawar Hayat$^1$}
\institute{$^1$ Monash University, $^2$ IIT Ropar, $^3$ University of Melbourne \\
\texttt{\{shreya.ghosh, munawar.hayat\}@monash.edu, abhinav@iitrpr.ac.in, jarrod.knibbe@unimelb.edu.au}}
\authorrunning{Ghosh et al.}

\maketitle

\begin{abstract}
Over the past few years, there has been an increasing interest to interpret gaze direction in an unconstrained environment with limited supervision. Owing to data curation and annotation issues, replicating gaze estimation method to other platforms, such as unconstrained outdoor or AR/VR, might lead to significant drop in performance due to insufficient availability of accurately annotated data for model training. In this paper, we explore an interesting yet challenging problem of gaze estimation method with a limited amount of labelled data. The proposed method distills knowledge from the labelled subset with visual features; including identity-specific appearance, gaze trajectory consistency and motion features. Given a gaze trajectory, the method utilizes label information of only the start and the end frames of a gaze sequence. An extension of the proposed method further reduces the requirement of labelled frames to only the start frame with a minor drop in the generated label's quality. We evaluate the proposed method on four benchmark datasets (CAVE, TabletGaze, MPII and Gaze360) as well as web-crawled YouTube videos. Our proposed method reduces the annotation effort to as low as 2.67\%, with minimal impact on performance; indicating the potential of our model enabling gaze estimation `in-the-wild' setup\footnote{https://github.com/i-am-shreya/Labelling-the-Gaps}.
\end{abstract}

\section{Introduction}
	
The `language of the eyes' provides an insight into a complex mental state such as visual attention~\cite{liu2011visual} and human cognition (emotions, beliefs and desires)~\cite{wang2017deep}. Accurate gaze estimation has wide applications in computer vision-related assistive technologies~\cite{mustafa2018prediction,ghosh2020speak2label} where the gaze is measured as a line of sight of the pupil in 3D/2D space or 2D screen location~\cite{gaze360_2019}. 

Recent advances in computer vision and deep learning have significantly enhanced the accuracy of gaze estimation~\cite{ghosh2021Automatic}. Most promising eye gaze estimation techniques either require specialized hardware (for example Tobii~\cite{niehorster2020glassesviewer}) or use supervised image processing solutions~\cite{zhang2017everyday,smith2013gaze}. Device and sensor based gaze estimation methods are highly dependent on user assistance, illumination specificity, the high device failure rate in an uncontrolled environment and constraints on the device's working distance. On the other hand, supervised methods require a large amount of labelled data for training. Manual labelling of human gaze information is a complex, noisy, resource-expensive and time-consuming task. To overcome these limitations, weakly-supervised learning provides a promising paradigm since it enables learning from a large amount of readily available non-annotated data. Few works~\cite{kothari2021weakly,ghosh2022mtgls,yu2019unsupervised,dubey2019unsupervised} explore in this direction to eliminate the data curation and annotation issue. However, these methods mostly investigate from a spatial analysis perspective. Human eye movement is a spatio-temporal, dynamic process which is either task-driven or involuntary action. Thus, it would be interesting to simplify the ballistic eye movement and curate large-scale training data for gaze representation learning.

To this end, we propose a weakly supervised eye gaze estimation framework. Our proposed technique reduces the requirement of a large number of annotated training samples. We show that the technique can also be used to facilitate the annotation process and reduce the bias in the data annotations. The proposed method requires the ground truth labels of start and end frames in a pre-defined gaze trajectory. We further refine this strategy where only the start frame's gaze annotation is required. Our proposed method significantly reduces the annotation effort which could be beneficial for annotating large-scale gaze datasets quickly. Moreover, it can be used in several applications such as immersive, augmented and virtual reality~\cite{swaminathan2018enabling,blattgerste2018advantages} (especially in Foveated Rending (FR)), animation industry~\cite{gumilar2021connecting,park2021talking} and social robotics~\cite{Alonso_Mart_n_2012,zabala2021modeling}, where unsupervised or weakly supervised calibration is highly desirable. In Foveated Rending (FR), gaze based interaction demands low latency gaze estimation to reduce energy consumption. To achieve this, the virtual environment displays high-quality images only from the user's point of view and blurs the peripheral region. Due to the subsequent delays in the frame-wise gaze estimation pipeline, the usage of FR is quite limited and mostly headpose direction is used to approximate the field of view~\cite{blattgerste2018advantages}. Our proposed method has the potential to bridge the gap and reduce energy consumption by interpolating the gaze trajectory of the user. Another potential application includes animation industry~\cite{gumilar2021connecting,park2021talking}. Given the start and end Point of Gaze (PoG) of a virtual avatar, our method can easily generate realistic labels for intermediate frames to display realistic facial gestures in the interaction environment~\cite{gumilar2021connecting,park2021talking}. Similarly, in social robotics, multi-modal gaze control strategies have been explored for guiding the robot's gaze. For example, an array of microphones has been utilized~\cite{Alonso_Mart_n_2012} to guide the gaze direction of a robot named Maggie. The other well-established methods include the usage of infrared laser and multimodal stimuli (e.g., visual, auditory and tactile) for modelling any known gaze trajectories~\cite{zabala2021modeling}. Our proposed methods could eliminate the aforementioned requirement of specialised hardware or pre-defined heuristics to navigate the environment. The \textbf{main contributions} of the paper are as follows:

\begin{enumerate}
\item We propose two weakly supervised neural networks (2-labels, 1-label) for gaze estimation. `2-labels' and `1-label' require the labels of two and one frames in a gaze sequence, respectively. 
\item We use task-specific information to bridge the gap between labelled and unlabelled samples. Our proposed method leverages facial appearance, relative motion, trajectory ordering and embedding consistency. This task-specific knowledge bridge the gap between labelled and unlabeled samples via learning.
\item We evaluate the performance of the proposed networks in two settings: 1) \textit{On benchmark datasets} (CAVE, TabletGaze, MPII and Gaze360) and 2) \textit{On unlabelled `in the wild' YouTube data} where ground truth annotation is not available. Additionally, we perform cross-dataset experiments to validate the generalizability of the framework. 

\item We also demonstrate the effectiveness of our proposed techniques by re-learning state-of-the-art eye gaze estimation methods with the labels generated by our method with very few prior annotations. The results indicate comparable performance for state-of-the-art frameworks (for example, 3.8 degrees by pictorial gaze~\cite{park2018deep} and 4 degrees with our 2-label technique).
\item We also validate our learning based interpolation method on unlabelled YouTube data where ground truth annotation is not available. Our experimental results suggest that this annotation method can be useful for extracting substantial training data for learning gaze estimation models.

\end{enumerate}
The rest of the paper structure is as follows: Section \ref{sec:prior} describes prior works in this area. Section \ref{sec:Preliminaries} describes preliminaries regarding eye movements, gaze trajectory and notations. Section \ref{sec:method} is about  the  proposed  methods. Experimental details and results are discussed in Section \ref{sec:experiments} and \ref{sec:results}.  Conclusion, limitations and future research directions are discussed in Section \ref{sec:conclusion}.

\section{Related Work}
\label{sec:prior}

\noindent \textbf{Gaze Estimation.} Recent advances in computer vision and deep learning techniques have significantly enhanced the gaze estimation performance~\cite{park2018deep,gaze360_2019}. A thorough analysis of gaze estimation literature is mentioned in a recent survey~\cite{ghosh2021Automatic}. Appearance based gaze estimation methods~\cite{lu2017appearance,zhang2015appearance,lu2015gaze} learn image to gaze mapping either via support vector regression~\cite{smith2013gaze} or deep learning methods~\cite{krafka2016eye,zhang2015appearance,zhang2017s,zhang2017mpiigaze,jyoti2018automatic,FischerECCV2018}. Among the deep learning based methods, supervised learning methods~\cite{zhang2015appearance,jyoti2018automatic,zhang2017mpiigaze,park2018deep} mostly encode appearance based gaze which require a large amount of annotated data. To overcome the limitation, few works explore gaze estimation with limited supervision such as `learning-by-synthesis' \cite{sugano2014learning}, hierarchical generative models~\cite{wang2018hierarchical}, conditional random field~\cite{benfold2011unsupervised}, unsupervised gaze target discovery~\cite{zhang2017everyday}, unsupervised representation learning~\cite{dubey2019unsupervised,yu2019unsupervised}, weakly supervised learning~\cite{kothari2021weakly}, pseudo labelling~\cite{ghosh2022mtgls} and few-shot learning~\cite{park2019few,yu2019improving}. Among these studies, the few shot learning approach required very few ($\leq 9$) calibration samples for gaze inference. Our method requires even less data annotation for gaze estimation (CAVE: 6.56\%, TabletGaze: $<1\%$, MPII: 4.67\% and Gaze360: 2.38\%).  

\noindent \textbf{Gaze Motion.} Eye movements are divided into the following categories: \textit{1) Saccade.} Saccades are voluntary eye movements to adjust the PoG in a visual field and it usually lasts for 10 to 100 ms. \textit{2) Smooth Pursuit.} It is an involuntary eye movement that occurs while tracking a moving visual target. \textit{3) Fixations.} Fixations consist of three involuntary eye movements termed as \textit{tremor, drift} and \textit{microsaccades}~\cite{duchowski2017eye}. The main objective is to stabilize the PoG on an object. Prior works along this line mainly used velocity based thresholding~\cite{komogortsev2013automated}, BLSTM~\cite{startsev20191d}, Bayesian framework~\cite{santini2016bayesian}, and hierarchical HMM~\cite{zhu2020hierarchical} for classification. Arabadzhiyska et al.~\cite{arabadzhiyska2017saccade} model saccade dynamics for gaze-contingent rendering. Our proposed method uses trajectory constrained gaze interpolation using temporal coherency with limited ground truth labels.

\begin{table}[b]
\caption{\small Comparison of benchmark datasets for cost analysis.} 
\label{tab:datasets_cost} 
\centering
\scalebox{0.85}{
\begin{tabular}{l||l}
\toprule[0.4mm]
\rowcolor{mygray}
\multicolumn{1}{l||}{\textbf{Dataset}} & \multicolumn{1}{l}{\textbf{Cost Analysis}}  \\ \hline \hline

\begin{tabular}[c]{@{}l@{}}CAVE~\cite{smith2013gaze}\end{tabular}  & \begin{tabular}[c]{@{}l@{}}  Canon EOS Rebel T3i camera and a Canon EF-S\\ 18–135 mm IS f/3.5–5.6 zoom lens  \end{tabular}                                                          \\ \hline

\begin{tabular}[c]{@{}l@{}}MPII~\cite{zhang2015appearance}\end{tabular}  & \begin{tabular}[c]{@{}l@{}} Laptop \\ Collection duration: 3 months\end{tabular}                                                                   \\ \hline

\begin{tabular}[c]{@{}l@{}}TabletGaze~\cite{huang2015tabletgaze}\end{tabular}  & \begin{tabular}[c]{@{}l@{}} Samsung Galaxy Tab S\end{tabular}                                                        \\ \hline
\begin{tabular}[c]{@{}l@{}}Gaze360~\cite{gaze360_2019}\end{tabular}  & \begin{tabular}[c]{@{}l@{}} Ladybug5 360\textdegree panoramic camera, AprilTag \end{tabular}                                                       \\ \hline
\begin{tabular}[c]{@{}l@{}}ETH-XGaze~\cite{zhang2020eth}\end{tabular}  & \begin{tabular}[c]{@{}l@{}} 18 Canon 250D SLR camera, ESPER trigger box,\\ Raspberry Pi and with controlled illumination.\end{tabular}                                                       \\ 
\bottomrule[0.4mm]
\end{tabular}}
\vspace{-8mm}
\end{table}

\begin{figure*}[t]
    \centering
    \includegraphics[width = \linewidth]{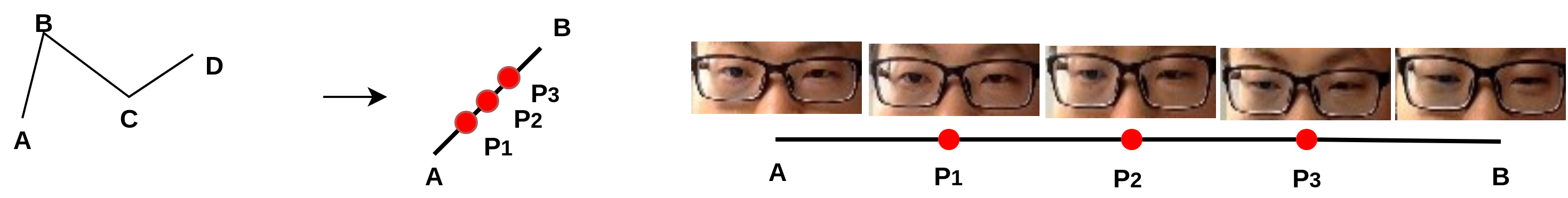}  
    \caption{\small Weakly supervised labelling approach illustration. The zig-zag path $\ora{AD}$ in the left is an example of a human gaze movement. This path can further be broken into several `gaze trajectories' ($\ora{AB}, \ora{BC}$ and $\ora{CD}$). Given a gaze trajectory $\ora{AB}$ with gaze annotation for A and B, the objective of this work is to annotate the unlabelled frames i.e. $P_1$, $P_2$ and $P_3$. The right side is an example of a gaze trajectory~\cite{zhang2015appearance}.}
    \label{fig:problem}
    \vspace{-5mm}
\end{figure*}

\noindent \textbf{Gaze Datasets.} In the past decade, several datasets~\cite{smith2013gaze,FunesMora_ETRA_2014,huang2015tabletgaze,zhang2017mpiigaze,zhang2017s,FischerECCV2018,gaze360_2019,garbin2019openeds,palmero2020openeds2020} have been proposed to estimate gaze accurately. The dataset collection technique has evolved from constrained lab environments~\cite{smith2013gaze,zhang2020eth} to unconstrained indoor~\cite{huang2015tabletgaze,zhang2017mpiigaze,zhang2017s,FischerECCV2018,zhang2020eth} and outdoor settings~\cite{gaze360_2019}. To consider both the constrained and unconstrained settings, we evaluate our weakly supervised framework in CAVE~\cite{smith2013gaze}, TabletGaze~\cite{huang2015tabletgaze}, MPII~\cite{zhang2017mpiigaze} and Gaze360~\cite{gaze360_2019} datasets. The data collection process requires either manual or sensor based annotations (Refer Table~\ref{tab:datasets_cost}). As this being time consuming process, the research community moves towards the data generation process for benchmarking with a large variation in data attributes. Prior works in this domain generate both synthetic and real image. In order to capture the possible rotational variation in image, gaze redirection techniques~\cite{ganin2016deepwarp,he2019photo,wood2018gazedirector,yu2019improving} are quite popular. The other approaches are mainly based on random forest~\cite{kononenko2015learning} and style transfer~\cite{sela2017gazegan}. However, due to several image quality based limitations, these generated datasets are not used for benchmarking.

\noindent \textbf{Weakly Supervised Neural Networks.}
Over the past few years, several promising weakly supervised methods have been proposed which mainly infer on the basis of prior knowledge~\cite{bilen2016weakly,lee2013pseudo}, task-specific domain knowledge~\cite{zhang2018weakly,arandjelovic2016netvlad,zhang2018bilateral}, representation learning~\cite{weston2012deep,zhao2015stacked}, loss-imposed learning paradigms~\cite{arandjelovic2016netvlad,sajjadi2016mutual} and combinations of the above~\cite{arandjelovic2016netvlad}.  Williams et al.~\cite{williams2006sparse} propose a semi-supervised Gaussian process model to predict the gaze. This method simplifies the data collection process as well. Bilen et al.~\cite{bilen2016weakly} use pre-trained deep CNN for the object detection task. Arandjelovic et al.~\cite{arandjelovic2016netvlad} propose a weakly supervised ranking loss for the place recognition task. Haeusser et al.~\cite{haeusser2017learning} introduce `associative learning' paradigm, which allows semi-supervised end-to-end training of any arbitrary network architecture. Unlike these studies, we explore loss-imposed domain knowledge for our framework.

\section{Preliminaries}
\label{sec:Preliminaries}

\noindent \textbf{Gaze Trajectory.} Human eye movement follows an arbitrary continuous path in three-dimensional space termed as the `gaze trajectory'~\cite{purves2015perception}. Gaze trajectories generally depend on the person, context and external factors~\cite{majaranta2011gaze}. Eye movements can be divided into three types: fixations, saccades and smooth pursuit~\cite{purves2015perception}. Fixation occurs when the gaze may pause in a specific position voluntarily or involuntarily. Conversely, gaze moves from one to another position for a saccade. The human gaze consists of a series of fixations and saccades in random order. In this work, we consider a small duration of eye movement from one position to another. Let us assume, the zig-zag path (left of Fig. \ref{fig:problem}) is an example of the human gaze movement path. This path can be divided into three small sub-paths ($\overrightarrow{AB}$, $\overrightarrow{BC}$ and $\overrightarrow{CD}$). We conduct our experiments on each such small sub-path. We use the term `gaze trajectory' to refer to this simplified version of the gaze path (i.e. $\overrightarrow{AB}$). The red points ($P_1,P_2$ and $P_3$) in $\overrightarrow{AB}$ are the frames in the A to B sequence. Considering these as discrete, they can be split into 3-point subsets with a constraint: each subset should contain the start and endpoints and the points should maintain the order of trajectory sequence. We term the 3 points set as \textit{3-frame set}. For example, $\{A,P_1,B\}$, $\{A,P_2,B\}$ and $\{A,P_3,B\}$ are three 3-frame sets.

\noindent \textbf{Problem Statement.}
In the context of a video, the points (A, $P_1$, $P_2$, $P_3$ and B) are frames in a specific trajectory order. Throughout this paper, we term A and B as start and end frames. Given the annotated start and end frames, the gaze trajectory sequence can be divided into small segments. There might be a high and insignificant temporal coherence if the segment duration is too small. On the other hand, a longer duration could affect the learning of meaningful representation due to diversity. We observed that the average difference in large time segment consisting of approx. $\sim 80$ frames is around $\sim 35$\textdegree. On the other hand, the smallest segment has angular difference of $<1$\textdegree. On this front, from a long sequence, we mine 3-frame subsets of gaze trajectory for learning meaningful representation. We work on two experimental settings: (a) \textbf{2-labels}: \textit{When the start and end frame annotations are available}. (b) \textbf{1-label}: \textit{When only the start frame annotation is available}. Given a gaze trajectory similar to $\overrightarrow{AB}$ with labels for A and B, the objective of this work is to annotate the unlabelled frames i.e. $P_1$, $P_2$, $P_3$, \dots $P_n$ where, n is the number of intermediate frames in a `gaze-trajectory'.

\noindent \textbf{Notations.}
Suppose that we have a set of N `3-frame set' samples in a dataset $D = \{X_{n}, Y_{s_{n}}, Y_{e_{n}}\}_{n=1}^N$, where $X_n$ is a $n^{th}$ 3-frame set consisting of start, middle/unlabelled and end frames $(f_s,f_{ul},f_e)_n$, $Y_{s_{n}}$ and $Y_{e_{n}}$ are the $n^{th}$ start and end frame labels, respectively. Lets assume our model $G$ with learnable parameter $\theta$ maps input $X_n \in \mathbb{R}^{3\times 100 \times 50 \times 3}$ to the relevant label spaces i.e., $Y_{s_{n}} \in \mathbb{R}^{3}$, $Y_{ul_{n}} \in \mathbb{R}^{3}$ and $Y_{e_{n}} \in \mathbb{R}^{3}$. The mapping function is denoted as  $G_{\theta}: X_n \to \{Y_{s_{n}},\ Y_{ul_{n}}, Y_{e_{n}}\}$.

\section{Gaze Labelling Framework}
\label{sec:method}

\subsection{Architectural Overview of `2-labels'}
\begin{figure*}[t]
    \centering
    \includegraphics[width = \linewidth]{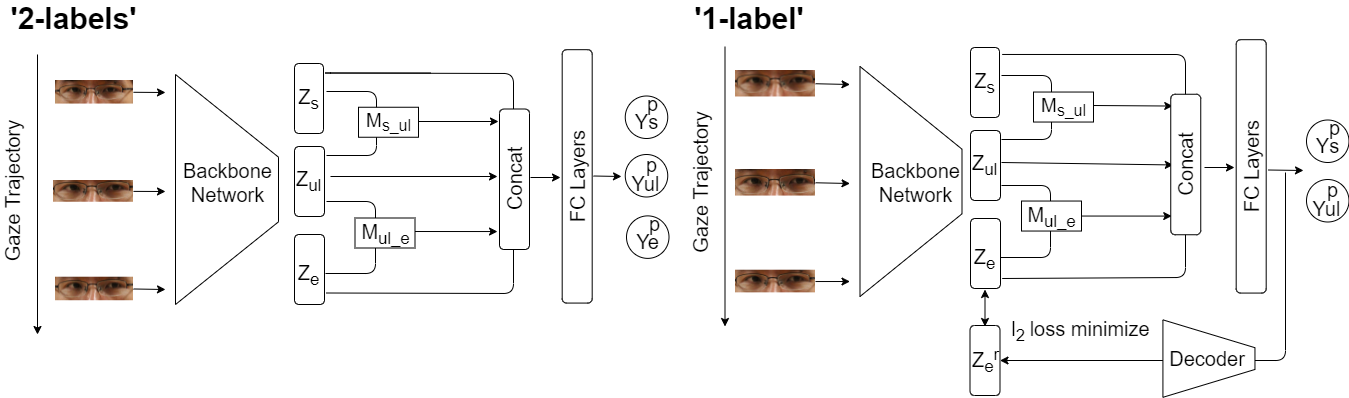}  
    \caption{\small Overview of `2-labels' and `1-label' network pipelines. The frameworks take 3-frame set as input and learn to interpolate intermediate labels via weak supervision. Refer Sec.~\ref{sec:method} for more details.}
    \label{fig:pipeline}
    \vspace{-5mm}
\end{figure*}

The overview of the proposed framework is shown in Fig.~\ref{fig:pipeline}. Given a 3-frame set $X_t:\{f_s,f_{ul},f_e\}$, we define an encoder $E$ which maps the input $X_t$ to latent space $Z_t:\{Z_s,Z_{ul},Z_e\}$, where $ Z_s\in \mathbb{R}^{2048},Z_{ul}\in \mathbb{R}^{2048},Z_e\in \mathbb{R}^{2048}$ (Refer Fig.~\ref{fig:pipeline} Left). After $E: X_t \rightarrow Z_t$ mapping,\ two motion features $M_{s\_ul}$ and $M_{ul\_e}$ are extracted between start-middle frames and middle-end frames. These motion features are concatenated with the latent embeddings. On top of it, Fully Connected (FC) layers having 512 and 1024 nodes are appended before prediction. Finally, the network predicts $Y_s^p$, $Y_{ul}^p$ and $Y_{e}^p$, where, $Y_s^p$, $Y_{ul}^p$ and $Y_{e}^p$ are gaze information corresponding to the input frames \{$f_s,f_{ul},f_e$\}. The backbone network is not architecture-specific, although we use VGG-16~\cite{simonyan2014very} and Resnet-50~\cite{he2016deep} for our experiments. The rationale behind the framework design is explained as follows:

\noindent \textbf{Identity Adaptation} The obvious usefulness of user adaptation of gaze calibration for AR and VR devices motivates us to design an identity specific gaze labelling framework~\cite{schmidt2020depth}. Thus, we purposefully select identity specific 3-frame sets. At the same time, it is important for the framework to be able to learn the variations across a large number of subjects with a different head pose, gaze direction, appearance, illumination, image resolution and many other configurations. Additionally, the framework should encode rich features relevant to eye region appearance, which is the most important factor for weakly supervised gaze labelling. Similar to recent studies~\cite{zhang2017s,park2019few}, we use eye region images as input for gaze inference. The 3-frame set is selected over a small duration temporal window in a video. As there is a subtle change in the appearance of eyes from one frame to another, the latent representation of the image also has minimal change. At a conceptual level, we are motivated by the smoothness constraint in optical flow algorithms. We choose to calculate cosine distance between start and middle, middle and end pair while calculating consistency loss. It helps in preserving the identity specific features across the 3-frame set.

\noindent \textbf{Motion Feature.} At first glance, the task of predicting gaze from sparsely labelled data may seem overly challenging. However, given a `3-frame set' sequence of a subject in very small time duration, there will be a high correspondence between frame A, $P_i$ and B, where $i=\{1,2,3\}$. Given this constraint, the objective is reduced to modelling the head and eye motion information to bridge the gap. This motivates us to use motion features for sequence modelling. Similar to~\cite{bertasius2019learning} in pose estimation domain, we encode a weak inter-frame motion by computing the $\ell_1$ distance between two consecutive frames in the latent space. We define motion feature by $M_{s\_ul} \oplus M_{ul\_e}$
where $M_{s\_ul} = Z_s -Z_{ul}$ and $M_{ul\_e} = Z_{ul} - Z_e$. $M_{s\_ul}$ and $M_{ul\_e}$ represent motion feature between start-middle and middle-end frames, respectively. Further, we use this feature to estimate the gaze-direction in a given trajectory. To train the above-mentioned network, we use the following loss functions.

\noindent \textbf{Regression Loss.} Corresponding to each 3-frame set $X_t$, the start and end frames are annotated in the 2-labels setting. These annotations provide strong supervisory information to predict the gaze information of the middle unlabelled frame. It provides information regarding an arbitrary gaze trajectory. The unlabelled middle frame lies in between start and end frames in that specific trajectory. Thus, it belongs to the same distribution of the start and end frames. The regression loss is defined as: $l_{reg} = MSE(Y_s,Y_s^p) + MSE(Y_e,Y_e^p)$
Here, MSE is Mean Squared Error, $Y_s$, $Y_s^p$, $Y_e$ and $Y_e^p$ are start label, predicted start label, end label and predicted end label, respectively.

\noindent \textbf{Consistency Loss.} The main aim of the consistency loss is to maintain consistency between latent and label space. Here, latent space is denoted as $Z$ and label space is the output space of the `2-labels' framework. In the latent space, let the distance between $Z_s$ and $Z_{ul}$ be $d_{s\_ul}^{z}$ and the distance between $Z_{ul}$ and $Z_e$ be $d_{ul\_e}^{z}$. Similarly in the label space, let the distance between $Y_s^p$ and $Y_s^p$ be $d_{s\_ul}^{y}$ and the distance between $Y_s^p$ and $Y_e$ be $d_{ul\_e}^{y}$. According to our hypothesis, the distance from start to unlabelled frame and unlabelled to end frame remains consistent. The loss is defined as follows: $l_{consistency} = \{|d_{s\_ul}^{y}-d_{s\_ul}^{z}|+|d_{ul\_e}^{y}-d_{ul\_e}^{z}|\}$

In the equation, the cosine distance is considered. The rationale behind the choice is as follows: the cosine distance is applied pairwise to utilize the partial annotations and bridge the gap between labelled and unlabelled frames. The distance has following properties: it leverages the identity specific information across the 3-frame set and it captures the motion-similarity information which indirectly encodes relative ordering of the frames. The angular distance ($d_{a\_b}$) between the frames a and b is defined as follows: $d_{a\_b} = \frac{a}{||a||_2}. \frac{b}{||b||_2}$
where, a and b are latent or label space embeddings of the start, unlabelled and end frames. `$(.)$' denotes the dot product. According to~\cite{barz2020deep}, this term is calculated in two stages. First, the latent embeddings are $L_2$-normalized (i.e. $\frac{a}{||a||_2}$), which maps the d-dimensional latent embedding to a unit  hyper-sphere, where the cosine similarity/distance and dot product are equivalent. As the human gaze follows a spatio-temporal trajectory~\cite{purves2015perception}, the 3-frame sets satisfy their relative trajectory ordering as well. The condition is satisfied by the consistency loss as it incorporates their relative position w.r.t. the middle unlabelled frame from label space to the latent space.

\noindent \textbf{Overall Loss Function.} The final loss function is defined below:
\begin{equation}
 Loss = \lambda_1 l_{reg} + \lambda_2 l_{consistency}
\label{eq:two_loss}
\end{equation}

It includes regression and consistency losses. Here, $\lambda_{1,2}$ are the regularization parameters. Further, we reduce the label requirement (by removing the requirement of the end frame's label) in the next framework with minimal impact in the performance.

\noindent \subsection{Architectural Overview of `1-label'}

The network architecture of 1-label network is shown in Fig. \ref{fig:pipeline} Right. Similar to the 2-labels network, the motion feature and identity specific appearance are leveraged for gaze estimation. The main motivation for moving from 2-labels to 1-label architecture is to leverage rich features with even lesser number of total annotations in a dataset. In `1-label' architecture, we have additional decoder module defined as $D(Z'_e;\phi_D)$ consisting of an FC layer. The decoder $D$, parameterized by $\phi_D$, maps the penultimate layer features (i.e. $Z'_e \in \mathbb{R}^{1024}$) to latent-embeddings \ $Z_e^r \in {R}^{2048}$. The rationale behind this remapping is adding constraints which can enhance the performance of the network by encoding meaningful representation from $f_e$. $l_2$ loss is computed between $Z_e$ and $Z_e^r$. The backbone network is not architecture-specific similar to the 2-labels architecture, although we use VGG-16~\cite{simonyan2014very} and Resnet-50~\cite{he2016deep} as backbone networks for our experiments. To train this network, we use the following loss functions: 

\noindent \textbf{Regression Loss.} Similar to 2-labels architecture, we compute the regression loss corresponding to the start label as follows: $l_{reg} = MSE(Y_s,Y_s^p) $
Here, MSE is mean squared error, $Y_s$ and $Y_s^p$ are start and predicted start label, respectively.

\noindent \textbf{Consistency Loss.}
Similar to 2-labels, in the 1-label technique, we add the consistency loss $l_{consistency}$. The loss is defined as follows: $l_{consistency} = \{|d_{s\_ul}^{y}-d_{s\_ul}^{z}|\}$
Here, $d_{s\_ul}^{y}$ and $d_{s\_ul}^{z}$ are distance between start and unlabelled frame in label and latent space, respectively.

\noindent \textbf{Similar Distribution.}
We leverage on the constraint that the gaze information belongs to similar distribution. Given a 3-frame set $X_t$, the output gaze should belong to a specific distribution and for that we compute Kullback-Leibler divergence (KL). To computes KL divergence loss between $Y_{true}$ and $Y_{pred}$, the following equation is followed: $loss = Y_{true} (log\frac{Y_{true}}{Y_{pred}})$. The objective function $l_{divergence}$ is the loss term to minimize the divergence between the gaze information of start and unlabeled frames, defined as follows: $l_{divergence} = (KL(Y_s,Y_{ul}))$

\noindent \textbf{Embedding Loss.} It is the $\ell_2$ loss computed between $Z_e$ and $Z_e^r$. This pattern is predicting future embeddings from prior knowledge.

\noindent \textbf{Overall Loss Function.}
The final loss function for `1-label' is an ensemble of regression, consistency, similar distribution and embedding losses, where, $\lambda_{1} \cdots \lambda_{4}$ are the regularization parameters.
\begin{equation}
\begin{split}
     Loss = \lambda_1 l_{reg} +\lambda_2 l_{consistency}+\lambda_3 l_{divergence} +\lambda_4 l_{embedding} 
\end{split}
\label{eq:one_loss}
\end{equation}

\section{Experiments}
\label{sec:experiments}
We perform experiments on 4 benchmark datasets (CAVE, TabletGaze, MPII and Gaze360) as well as `in-the-wild' YouTube data, where the ground truth labels are not available. To validate the performance of the YouTube data, we generate pseudo labels (via SLERP and other methods) for comparison. The details are as follows:

\subsection{Automatic 3-frame set mining}
The first step to implement our proposed method is 3-frame set mining. For generating 3-frame sets, we perform dataset-specific pre-processing as the datasets are collected in different setups. Please note that we require ground truth labels to define the gaze trajectories, especially for CAVE and MPII datasets as temporal information is not present. Moreover, the annotated subset required for weak supervision is a special subset which contains start and end frame of the trajectories. Additionally, there is no temporal overlap between 3-frame sets for any of the datasets. Table~\ref{tab:data_amount} shows the comparison among dataset statistics to show the amount of images in the original data and derived data (3-frames set mined used for weak-supervision). 

\subsubsection{Automatic 3-frame set mining on benchmark datasets.}
We validate the proposed methods on 4 benchmark datasets: CAVE, Tabletgaze, MPII and Gaze360. These datasets are collected in fully constraint (CAVE), less constrained (Tabletgaze and MPII) and unconstrained (Gaze360) environments.

\begin{figure*}[t]
\includegraphics[width = 1.0\linewidth]{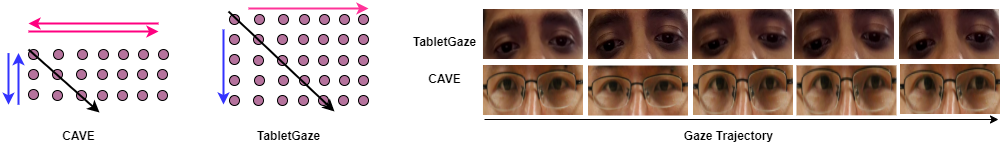} 
\caption{3-frame set mining process for CAVE~\cite{smith2013gaze} and TabletGaze dataset~\cite{huang2015tabletgaze}. Here, red, blue and black arrows represent horizontal, vertical and diagonal gaze trajectories.}
\label{fig:triplet_mining}
\vspace{-5mm}
\end{figure*}

\noindent \textbf{CAVE}~\cite{smith2013gaze} contains 5,880 images of 56 subjects with different gaze directions and head poses. There are 21 different gaze directions for each person and the data was collected in a constrained lab environment. CAVE dataset is collected with 7 horizontal and 3 vertical gaze locations as shown in the left part of Fig. \ref{fig:triplet_mining}. Considering these positions as $7\times 3$ grids, we defined three types of gaze trajectories: horizontal, vertical and diagonal. As temporal information is missing, we can consider bi-directional gaze trajectories. We reverse the order for bidirectional set mining (Refer Fig.~\ref{fig:triplet_mining}). The bidirectional gaze trajectories are applied for CAVE dataset only due to the absence of temporal information. Note that this does not impact the requirement of ground truth annotation for weak supervision. In this way, we collect 3,024 3-frame sets for training. For this dataset, we require 6.56\% and 3.28\% of prior data annotation for our `2-labels' and `1-label' paradigms. 

\noindent \textbf{TabletGaze}~\cite{huang2015tabletgaze} is a large unconstrained gaze dataset of 51 subjects with 4 different postures and 35 gaze locations collected using a tablet in an indoor environment. TabletGaze dataset is also collected in a $7\times5$ grid format (as shown in the middle image of Fig. \ref{fig:triplet_mining}). Similar to CAVE dataset, we define horizontal, vertical and diagonal gaze trajectories and collect 108,524 3-frame sets. For TabletGaze dataset, as temporal information is also present, we consider unidirectional frames only. For 3-frame set mining on TabletGaze data, we require less than 1\% prior data annotation for both the frameworks.

\noindent \textbf{MPII}~\cite{zhang2017mpiigaze} gaze dataset contains 213,659 images collected from 15 subjects during natural everyday events in front of a laptop over a three-month duration. MPII gaze dataset is collected by showing random points on the laptop screen to the participants. To make the gaze trajectory smooth, we sort the given coordinates of the points in ascending order and consider it as a gaze trajectory. Further, 3-frame sets are collected in a day-wise for each participant. Following this procedure, we collect 32,751 3-frame sets. We extracted these 3-frame sets with 4.67\% prior data annotation.

\noindent \textbf{Gaze360}~\cite{gaze360_2019} is a large-scale gaze estimation dataset collected from 238 subjects in unconstrained indoor and outdoor settings with a wide range of head pose. We compute person-specific 3-frame sets. As each participant fixates gaze at a moving target, we consider the target's trajectory as the gaze trajectory. This results in 197,588 3-frame sets and we use 2.38\% of annotated data. 

\subsubsection{3-frame set mining on unlabelled `in the wild' YouTube data.}

We evaluate our method on an `in the wild' data i.e. when the expert/ground truth labels are not available. We leverage two eye symmetry property i.e. the change in relative position of the iris is symmetrical while scanning 3D space~\cite{dubey2019unsupervised}. We collect approximately 400,000 frames from YouTube videos using this strategy mentioned below. 

\noindent \textbf{Gaze Trajectory Selection} 
As the relative positions of pupil-centers provide the most important information regarding gaze direction, we utilize this property to detect gaze trajectories. We utilize two eye symmetry property i.e. the change in relative position of the iris is symmetrical while scanning 3D space~\cite{dubey2019unsupervised}. Based on this hypothesis, we compare the vertical angles formed with the following points i.e. pupil-center, nose in both eyes. In Fig. \ref{fig:failure}, $I_1$ and $I_2$ are the pupil centers; V is the vertical direction w.r.t. the nose tip point and $\theta_1$, $\theta_2$ are the above mentioned angles. The change in $\theta_1$ and $\theta_2$ depicts the path of the gaze trajectory sequence. For example, if a person shifts his/her gaze from left to right, the values of the angles will be as follows: initially, $\theta_1$ will be greater than $\theta_2$; then gradually $\theta_1$ will decrease and $\theta_2$ will increase; finally, $\theta_2$ will be greater than $\theta_1$. Thus, by monitoring these angles we can approximate gaze trajectories. The heuristic also considers the trajectory segment if it starts from the middle until there is a change in any of the angles $\theta_1$ and $\theta_2$. Although the proposed method is robust to head movements within the range of −10\textdegree\ to 10\textdegree. After identifying the gaze trajectories, we annotate the start and end frames with OpenFace~\cite{baltruvsaitis2016openface} and collect possible 3-frame sets. In this 3-frame set collection and data annotation procedure, we require 5.34\% and 2.67\% annotation of the overall data for `2-labels' and `1-label' settings. According to literature~\cite{eberly2011fast}, human gaze trajectories are considered to be spherical. Thus for ground truth annotation, we label remaining frames by SLERP interpolation method~\cite{eberly2011fast}. Further, we apply our `2-label' and `1-label' model on this data.

\begin{table}[t]
\caption{\small Comparison of benchmark dataset statistics to show the amount of images in the original data and derived data (3-frames set mined).} 
\label{tab:data_amount} 
\centering
\scalebox{0.85}{
\begin{tabular}{l||c|c}
\toprule[0.4mm]
\rowcolor{mygray}
\multicolumn{1}{l||}{\textbf{Dataset}} & \multicolumn{1}{c|}{\textbf{Original Dataset}} & \multicolumn{1}{c}{\textbf{Derived Dataset}} \\ \hline \hline

\begin{tabular}[c]{@{}l@{}}CAVE~\cite{smith2013gaze}\end{tabular}  & \begin{tabular}[c]{@{}l@{}} 5,880  \end{tabular}   & \begin{tabular}[c]{@{}l@{}} 3,024  \end{tabular}                                                       \\ \hline

\begin{tabular}[c]{@{}l@{}}MPII~\cite{zhang2015appearance}\end{tabular}  & \begin{tabular}[c]{@{}l@{}} 213,659  \end{tabular}& \begin{tabular}[c]{@{}l@{}} 32,751  \end{tabular}                                                                   \\ \hline

\begin{tabular}[c]{@{}l@{}}TabletGaze~\cite{huang2015tabletgaze}\end{tabular}  & \begin{tabular}[c]{@{}l@{}}    1,428 min\end{tabular}& \begin{tabular}[c]{@{}l@{}} 108,524  \end{tabular}                                                        \\ \hline
\begin{tabular}[c]{@{}l@{}}Gaze360~\cite{gaze360_2019}\end{tabular}  & \begin{tabular}[c]{@{}l@{}} 172,000  \end{tabular}& \begin{tabular}[c]{@{}l@{}}  197,588 \end{tabular}                                                       \\ 
\bottomrule[0.4mm]
\end{tabular}}
\vspace{-5mm}
\end{table}

\subsection{Experimental Settings.}
We define the following terminologies for easy navigation in the upcoming sections. \textit{1) Original Data (OD):} It refers to benchmark dataset's unaltered data; \textit{2) Derived Data (DD):} 3-frame set mined data derived from OD; \textit{3) Original Labels (OL):} Original ground-truth labels provided with OD; and \textit{4) Predicted Labels (PL):} Labels predicted from `2-labels' and `1-label' methods (i.e. $Y^p_{ul}$). We perform experiments with the following settings:

\noindent \textbf{1) Validation w.r.t. Ground Truth Labels.} First, we applied `3-frame set mining' to obtain the DD (Refer Table~\ref{tab:data_amount}) from OD. Further, we split the DD into 80\%-20\% train-test splits without any identity overlap. We evaluate our proposed method with OL. 

\noindent \textbf{2) Label Quality Assessment via State-of-the-art Methods' Performance.} We train the state-of-the-art methods~\cite{park2018deep,huang2015tabletgaze,gaze360_2019} on PL and validate on OL for label quality assessment. 

\noindent \textbf{3) Experiments with Different Data Partitions.} We train state-of-the-art models~\cite{park2018deep,huang2015tabletgaze,gaze360_2019} with different input data settings as follows: a) start and end frames of 3-frame sets, b) 50\% of the whole data and c) newly labelled frames.

\noindent \textbf{4) Ablation Studies.} We have conducted extensive ablation studies to show the importance of loss function, motion feature, sequential modelling and regularization parameters.

\noindent \textbf{5) Gaze Labelling `in the wild'.} We collect `in the wild' gaze data from YouTube videos having \textit{creative common licence} and compare the label quality with various model based techniques~\cite{valenti2011combining,yamazoe2008remote,ishikawa2004passive}. 

\noindent Further, we did perform additional experiments to evaluate the generalizibity of the proposed method. 

\subsection{Evaluation Metrics}
For quantitative evaluation, we use Mean Absolute Error (MAE), Correlation Coefficient (CC) and Angular Error. Mean Absolute Error is calculated as: $\frac{\sum_{i=1}^{n} |y^p-y|}{n}$
Correlation coefficient is calculated as: $\frac{\sum_{i=1}^{n} (y_i-\overline{y})(y^{p}_{i}-\overline{y^p})}{\sqrt{\sum_{i=1}^{n} (y_i-\overline{y})^2\ \sum_{i=1}^{n}(y^{p}_{i}-\overline{y^p})^2}}$ Here, $y^{p}$ is the predicted label and $y$ is the ground truth label in normalized space, $(\overline{.})$ indicates mean across the samples. Similar to the previous methods~\cite{zhang2020eth,park2018learning,park2019few}, angular error is the average error across test data measured in terms of cosine angle between ground truth and predicted gaze direction. It is measured as follows:
$\frac{g}{||g||_2}. \frac{g'}{||g'||_2}$
Here, $g$ and $g'$ respectively denote the ground truth and predicted gaze in terms of 3D gaze direction vector.

\subsection{Training Details.}
After the 3-frame set mining, we apply Dlib face detector~\cite{sharma2016farec} for eye detection. If face detection (dlib) fails especially for Gaze360 dataset, we use cropped headpose provided with the dataset\footnote{https://github.com/erkil1452/gaze360/tree/master/dataset}. Otherwise, we use the resized input image. For the backbone network, we choose VGG-16~\cite{simonyan2014very} and ResNet-50~\cite{he2016deep} architectures. For training, we use SGD optimizer with 0.001 learning rate with $1\times e^{-6}$ decay per epoch. The values of $\lambda_1 \cdots \lambda_4$ are 1. In each case, the models are trained for 1,000 epochs with batch size 32 and early stopping.

\begin{table*}[t]
\caption{\small Comparison of `2-labels' and `1-label' techniques using MAE, CC and Angular Error. Both frameworks with both VGG-16 and ResNet-50 backbone are trained on 80\% of the Derived Data (DD, i.e. 3-frame set mined) and validated on 20\% of the DD. This 80\%-20\% partition does not have any identity overlap. Here, MAE: Mean Absolute Error, CC: Correlation Coefficient, TG: TabletGaze. }
\label{tab:results}
\centering
\scalebox{0.85}{
\begin{tabular}{c||c|c|c|c|c|c|c|c|c|c}
\toprule[0.4mm]
\multirow{3}{*}{\cellcolor{black!10}} & \multicolumn{4}{c|}{\cellcolor{black!10}\textbf{VGG-16}}                                                & \multicolumn{6}{c}{\cellcolor{black!10}\textbf{ResNet-50}}                                             \\ \cline{2-11} 
\cellcolor{black!10}\textbf{Dataset}& \multicolumn{2}{c|}{\cellcolor{black!10}\textbf{2-labels}} & \multicolumn{2}{c|}{\cellcolor{black!10}\textbf{1-label}} & \multicolumn{2}{c|}{\cellcolor{black!10}\textbf{2-labels}} & \multicolumn{2}{c|}{\cellcolor{black!10}\textbf{1-label}} & \cellcolor{black!10}\textbf{2-labels} & \cellcolor{black!10}\textbf{1-label}\\ \cline{2-11}
\cellcolor{black!10}& \cellcolor{black!10}\textbf{MAE}        & \cellcolor{black!10}\textbf{CC}        & \cellcolor{black!10}\textbf{MAE}        & \cellcolor{black!10}\textbf{CC}        & \cellcolor{black!10}\textbf{MAE}        & \cellcolor{black!10}\textbf{CC}        & \cellcolor{black!10}\textbf{MAE}        & \cellcolor{black!10}\textbf{CC}& \multicolumn{2}{c}{\cellcolor{black!10}\begin{tabular}[c]{@{}c@{}}\textbf{Angular Error} \\ (in \textdegree, TG: in cm)\end{tabular}}        \\ \hline \hline
\textbf{CAVE}                     & 0.29              & 0.85               & 0.53              & 0.21                  & 0.25              &    0.90                & 0.43              &   0.25              &3.05 &3.30   \\ 
\textbf{MPII}                     & 0.43              & 0.62               & 0.57              &   0.25                 & 0.42              &  0.63                  & 0.57              &0.25             &5.00  &  5.40     \\ 
\textbf{Gaze360}                  &  0.38            &   0.69                &   0.45            &   0.42                 &   0.34            & 0.71                   & 0.40             &  0.70          & 15.00 & 15.80    \\ 
\textbf{TabletGaze}               & 0.49              & 0.54               & 0.50              & 0.58                   & 0.47              &    0.55                & 0.49              &    0.55             & 2.27 & 2.61   \\ 
\textbf{\begin{tabular}[c]{@{}c@{}}YouTube\\`in the wild' \end{tabular}}               & 0.27              & 0.90               & 0.36              & 0.81                   & 0.24              &    0.92                & 0.34              &    0.86              & 9.41 &  12.07 \\ 
\bottomrule[0.4mm]
\end{tabular}}
\vspace{-5mm}
\end{table*}

\begin{table}[t]
\caption{\small Results of the re-trained state-of-the-art methods on MPII and CAVE dataset in terms of angular error in \textdegree. OP=Original Predictions is the result as mentioned in the original papers~\cite{park2018deep}.}
\label{tab:sota}
\centering
\scalebox{0.85}{
\begin{tabular}{c||c|c|c|c}
\toprule[0.4mm]
\cellcolor{black!10}\textbf{Dataset}& \cellcolor{black!10}\textbf{Method}          & \cellcolor{black!10}\textbf{Alexnet} & \cellcolor{black!10} \textbf{VGG 16} & \cellcolor{black!10}\begin{tabular}[c]{@{}c@{}}\textbf{Pictorial Gaze}\\ \textbf{\cite{park2018deep}} \end{tabular} \\ \hline \hline
\multirow{3}{*}{\begin{turn}{90}CAVE\end{turn}} &\textbf{OP} & 4.20              & 3.90             & 3.80                   \\ 
&\textbf{2-labels}    & 4.10             & 4.46            & 4.00                   \\ 
&\textbf{1-label}     & 4.55              & 4.84            & 4.36                   \\ \hline
\multirow{3}{*}{\begin{turn}{90}MPII\end{turn}}&\textbf{OP} & 5.70              & 5.40            & 4.50                              \\ 
&\textbf{2-labels}    & 5.90             & 5.79           & 4.70                               \\ 
&\textbf{1-label}     & 6.30              & 6.20            & 4.90                   \\ \hline
\multirow{2}{*}{Train:YouTube} & \textbf{CAVE} & -- & -- & 3.90  \\ 
 & \textbf{MPII} & -- & -- & 4.57  \\ 
\bottomrule[0.4mm]
\end{tabular}}
\vspace{-5mm}
\end{table}

\begin{table}[t]
\caption{\small Results of the re-trained methods on TabletGaze and Gaze360 dataset in terms of angular error in \textdegree and cm. OP=Original Predictions is the result as mentioned in the~\cite{gaze360_2019,huang2015tabletgaze}.}
\label{tab:sota_1}
\begin{minipage}[b]{\hsize}
\centering
\scalebox{0.85}{
\begin{tabular}{c||c}
\toprule[0.4mm]
\cellcolor{black!10}\textbf{TabletGaze}                                       & \cellcolor{black!10}\textbf{\begin{tabular}[c]{@{}c@{}}mHOG+SVR\\ {\cite{huang2015tabletgaze}}\end{tabular}} \\ \hline \hline
\textbf{OP} & 2.50                                                               \\ 
\textbf{2-labels}                                         & 2.70                                                               \\ 
\textbf{1-label}                                          & 3.10                                                               \\ 
\begin{tabular}[c]{@{}c@{}} Train:YouTube\end{tabular} & 2.30 \\ 
\bottomrule[0.4mm]
\end{tabular}
\hspace{3mm}
\begin{tabular}{c||c}
\toprule[0.4mm]
\cellcolor{black!10}\textbf{Gaze360}  & \cellcolor{black!10}\textbf{\begin{tabular}[c]{@{}c@{}}Pinball LSTM\\~\cite{gaze360_2019}\end{tabular}} \\ \hline \hline
\textbf{OP}   & 13.50                                                                  \\ 
\textbf{2-labels} & 14.40                                                                  \\ 
\textbf{1-label}  & 17.20                                                                  \\ 
\begin{tabular}[c]{@{}c@{}} Train:YouTube\end{tabular} & 12.80 \\ 
\bottomrule[0.4mm]
\end{tabular}}
\end{minipage}
\vspace{-5mm}
\end{table}

\begin{SCtable}[][htp]
\centering
\caption{\small Performance comparison for different input data settings with the state-of-the-art methods on MPII, CAVE, TabletGaze (TG) and Gaze360 dataset. NL: Newly Labelled, SF: Start Frame, EF: End Frame.}
\label{tab:baselines}
\scalebox{0.85}{
\begin{tabular}{l||c|c|c|c}
\toprule[0.4mm]
\cellcolor{black!10}\textbf{Dataset}   &  \cellcolor{black!10}\textbf{CAVE}         & \cellcolor{black!10}  \textbf{TG} & \cellcolor{black!10} \textbf{Gaze360}     & \cellcolor{black!10}  \textbf{MPII}     \\ \hline
\cellcolor{black!10} \textbf{Method}   & \cellcolor{black!10} \textbf{\cite{park2018deep}} & \cellcolor{black!10} \textbf{\cite{huang2015tabletgaze}}  & \cellcolor{black!10}  \textbf{\cite{gaze360_2019}} & \cellcolor{black!10} \textbf{\cite{park2018deep}} \\ \hline \hline
(SF+EF)            & 7.14                    &  8.60           &  22.10                 &   6.10                  \\ 
50\% data           &   5.50                 & 4.50                 & 18.70                    &  5.40                    \\ 
2-labels & 4.00                  & 2.70             & 14.40                  & 4.70                   \\ 
NL frames     & 4.30                    & 3.00                 & 14.90                    &   4.30                   \\ 
1-label  & 4.36                   & 3.10                & 17.20                  &  4.90                   \\
\bottomrule[0.4mm]
\end{tabular}}
\vspace{-3mm}
\end{SCtable}

\begin{SCtable}[][htp]
\centering
\caption{\small Ablation study: effect of loss functions for `2-labels' (Eq. \ref{eq:two_loss}) \& 1-label (Eq. \ref{eq:one_loss}). NA: Not Applicable}
\label{tab:ablation_loss}
\scalebox{0.85}{
\begin{tabular}{l||c|c}
\toprule[0.4mm]
\cellcolor{black!10}\textbf{Loss}                                          & \cellcolor{black!10}\textbf{\begin{tabular}[c]{@{}c@{}}MAE \\ (2-labels)\end{tabular}} & \cellcolor{black!10}\textbf{\begin{tabular}[c]{@{}l@{}}MAE \\ (1-label)\end{tabular}} \\ \hline \hline
$\mathbf{l_{reg}}$                                                 & 0.98                                                           & 0.99                                                           \\ \hline 
$\mathbf{l_{reg}+l_{consistency}}$                                   & 0.42                                                           & 0.76                                                          \\ \hline
\begin{tabular}[c]{@{}c@{}}$\mathbf{l_{reg}+l_{consistency}+l_{divergence}}$ \\(for 1-label)
\end{tabular}& NA & 0.58                                                           \\ \hline
\begin{tabular}[c]{@{}c@{}}$\mathbf{l_{reg}+l_{consistency}+l_{divergence}}$\\$\mathbf{+l_{embedding}}$ (for 1-label) \end{tabular} & NA & 0.57  \\ 
\bottomrule[0.4mm]
\end{tabular}}
\vspace{-5mm}
\end{SCtable}

\section{Results}
\label{sec:results}
 
\subsection{Gaze Labelling and Estimation Performance Comparison.} 
To show the effectiveness of the proposed method, we evaluate on four benchmark datasets and YouTube data. First, we applied 3-frame set mining to get the derived data (Refer Table~\ref{tab:data_amount}). Further, we split the derived data randomly into train and test sets (train set: 80\% and test set: 20\%). Please note that the test set does not have identity overlap with training partition. The results are mentioned in Table \ref{tab:results} in terms of MAE, CC and angular error. We use VGG-16 and Resnet-50 as backbone networks to show the impact of different network architectures. Quantitatively, ResNet-50 performs slightly better than the VGG-16. From Table \ref{tab:results}, it is also observed that `2-labels' technique is closer to the original label distribution as compared to the `1-label' technique due to the absence of supervisory signal (i.e. absence of end frame label). Due to high-resolution images in CAVE dataset, generated labels' similarity as compared with original labels is high for `2-labels' setting. Instead of sequential modelling, our loss imposed gaze estimation method improves model performance.

\subsection{Cross Dataset Evaluation}\label{sec:cross_data}
We perform a cross dataset evaluation for predicting the generalization ability of the proposed method. This evaluation is conducted in two settings: The first configuration is a classical cross dataset evaluation protocol. In the second configuration, the training is performed on the train partition of OD using the PL i.e. output of `2-labels' technique $Y_{ul}^p$ and it is evaluated on the test partition of other datasets. The second configuration is conducted for cross dataset label quality assessment of the proposed method.

The first configuration is a classical cross dataset evaluation. Given any two datasets $D_1$ and $D_2$, the training is performed on the train partition of $D_1$ (i.e. $D_1^{train}$) using the OL and ResNet-50 as backbone network. Further, it is evaluated on test partition of $D_2$ (i.e. $D_2^{test}$). The results are shown in Table \ref{tab:cross_data_original}. In the second configuration, the training is performed on the train partition of $D_1$ (i.e. $D_1^{train}$) using the PL i.e. output of `2-labels' technique $Y_{ul}^p$, while it is evaluated on the test partition of $D_2$ (i.e. $D_2^{test}$). The results are depicted on Table~\ref{tab:cross_data_predicted}.
Please note that CAVE, TabletGaze and MPII do not have proper train-validation-test partitions~\cite{smith2013gaze,zhang2017mpiigaze,huang2015tabletgaze}. We train the model on $D_1$ and evaluate on $D_2$. From both the Tables \ref{tab:cross_data_original} and \ref{tab:cross_data_predicted}, it is observed that with the predicted label, the cross dataset performance is increased significantly which in turn, shows the generalization capability of our models. Apart from generalizability, it is also observed that the `1-label' framework performs comparatively well even after it is trained with less supervision. 

For `in-the-wild' data, the cross dataset performance is depicted in Table \ref{tab:cross_data_predicted} for both 2-labels and 1-label frameworks. Here (Table~\ref{tab:cross_data_predicted}), we have not fine-tuned the model. 

\begin{table*}[!htbp]
\centering
\caption{\small Cross dataset performance evaluation among the benchmark datasets \textit{in terms of MAE} in both `2-labels' and `1-label' settings. Given any two datasets $D_1$ and $D_2$, the training is performed on the train partition of $D_1$ (i.e. $D_1^{train}$) using the Original Label (OL) and ResNet-50 as backbone network. Further, it is evaluated on test partition of $D_2$ (i.e. $D_2^{test}$).}
\label{tab:cross_data_original}
\scalebox{1.0}{
\begin{tabular}{|c|c||c|c|c|c|}
\toprule[0.4mm]
\multirow{5}{*}{\begin{turn}{90}Original Label\end{turn}} & \cellcolor{black!10}\textbf{\begin{tabular}[c]{@{}c@{}}2-labels\\ Test$\rightarrow$\\ Train\\$ \downarrow$\end{tabular}} & \cellcolor{black!10}\textbf{CAVE} & \cellcolor{black!10}\textbf{MPII} & \cellcolor{black!10}\textbf{Gaze360} & \cellcolor{black!10}\textbf{TabletGaze} \\  \cline{2-6}\noalign{\vskip\doublerulesep \vskip-\arrayrulewidth} \cline{2-6}
                                         & \textbf{CAVE}                                                            & --            & 0.50        & 0.55          & 0.49                          \\  
                                         & \textbf{MPII}                                                            & 0.35       & --            &  0.53         & 0.51             \\ 
                                         & \textbf{Gaze360}                                                         & 0.21      &    0.40     & --               &    0.44          \\  
                                         & \textbf{TabletGaze}                                                      & 0.36       & 0.42       &  0.50         & --     \\ 
\bottomrule[0.4mm]
\end{tabular}}\\
\begin{tabular}{|c|c||c|c|c|c|}
\toprule[0.4mm]
\multirow{5}{*}{\begin{turn}{90}Original Label\end{turn}} & \cellcolor{black!10}\textbf{\begin{tabular}[c]{@{}c@{}}1-label\\ Test$\rightarrow$\\ Train\\$ \downarrow$\end{tabular}} &\cellcolor{black!10} \textbf{CAVE} &\cellcolor{black!10} \textbf{MPII} & \cellcolor{black!10}\textbf{Gaze360} & \cellcolor{black!10}\textbf{TabletGaze}\\ \cline{2-6}\noalign{\vskip\doublerulesep \vskip-\arrayrulewidth} \cline{2-6}
                                         & \textbf{CAVE}                                                           & --            & 0.54        & 0.60            & 0.50           \\  
                                         &  \textbf{MPII}                                                           & 0.57        & --            &   0.61         & 0.90                            \\ 
                                         &        \textbf{Gaze360}                                                        & 0.24        & 0.39        & --               & 0.50       \\  
                                         &  \textbf{TabletGaze}                                                     & 0.27        & 0.49        &    0.59        & --                            \\ 
\bottomrule[0.4mm]
\end{tabular}
\vspace{-5mm}
\end{table*}

\noindent \textbf{Comparison with State-of-the-art Methods.} 
We also evaluate the label generation quality of our proposed methods. For this purpose, we conduct experiments by training existing state-of-the-art methods~\cite{park2018deep,gaze360_2019,FischerECCV2018,huang2015tabletgaze} with the labels predicted from our method. The state-of-the-art network's performance is measured by comparing with the original ground truth labels. The performance comparison is mentioned in the Table \ref{tab:sota} and \ref{tab:sota_1}. For~\cite{park2018deep,gaze360_2019}, we use author's GitHub implementations~\cite{github_gaze}. It is observed that `2-labels' performs better than `1-label' for all the datasets. For CAVE dataset, `gazemap' based method~\cite{park2018deep} for `2-labels' (i.e. 4.08\textdegree) performs better than other settings. It is to be noted that the results of re-trained methods are comparable to when they were trained with the original labels. By using less than 5\% labelled data (For MPII 4.67\% and Gaze360 2.38\%), the labels generated by our weakly-supervised method perform favourably when evaluated on state-of-the-art methods~\cite{park2018deep,gaze360_2019}. This shows the usefulness of our weakly-supervised approach of label generation.

\noindent \textbf{Re-train State-of-the-art Methods with Subset of Data.}
To further validate the generated labels, we perform following experiments: we re-train from scratch~\cite{park2018deep},~\cite{huang2015tabletgaze} and~\cite{gaze360_2019} using the following labelled sets independently: a) start and end frames of 3-frame sets only, b) 50\% of originally labelled training data, c) frames with labels generated with 2-labels method, d)  frames with labels generated with 2-labels method apart from start and end frames, and e)  frames with labels generated with 1-labels method. The results are shown in Table \ref{tab:baselines}. When we train the networks with start and end frames, the error is high as the start and end frames consist of $<10\%$ of the whole dataset. When we use 50\% of the whole labelled data, the results significantly improve. Similarly, for the newly labelled frames, the error is less as compared to the above two settings. Please note that in the newly labelled case the training is performed on 90-95\% of the training data. These results validate the quality of labels generated by our methods. 

\begin{table*}[!htbp]
\centering
\caption{\small Cross dataset performance evaluation among the benchmark datasets and YouTube data \textit{in terms of MAE} in both `2-labels' and `1-label' settings. Given any two datasets $D_1$ and $D_2$, the training is performed on the train partition of $D_1$ (i.e. $D_1^{train}$) using the Predicted Label (PL) i.e. output of `2-labels' technique $Y_{ul}^p$. Further, it is evaluated on the test partition of $D_2$ (i.e. $D_2^{test}$).}
\label{tab:cross_data_predicted}
\scalebox{1.0}{
\begin{tabular}{|c|c||c|c|c|c|}
\toprule[0.4mm]
\multirow{6}{*}{\begin{turn}{90}Predicted Label\end{turn}} & \cellcolor{black!10}\textbf{\begin{tabular}[c]{@{}c@{}}2-labels\\ Test$\rightarrow$\\ Train\\$ \downarrow$\end{tabular}} & \cellcolor{black!10}\textbf{CAVE} & \cellcolor{black!10}\textbf{MPII} & \cellcolor{black!10}\textbf{Gaze360} & \cellcolor{black!10}\textbf{TabletGaze} \\ \cline{2-6} \noalign{\vskip\doublerulesep \vskip-\arrayrulewidth} \cline{2-6}
                                          & \textbf{CAVE}                                                            & --            & 0.54        &  0.51        & 0.47  \\ 
                                          & \textbf{MPII}                                                            & 0.27      & --            &  0.50         & 0.48     \\ 
                                          & \textbf{Gaze360}                                                         &   0.20     &  0.34       & --               &    0.42      \\  
                                          & \textbf{TabletGaze}                                                      & 0.32        & 0.44        & 0.50           & --                   \\ 
                                          & \textbf{YouTube}                             & 0.25        & 0.45         &  0.46         & 0.43                                        \\ 
\bottomrule[0.4mm]
\end{tabular}}\\
\begin{tabular}{|c|c||c|c|c|c|}
\toprule[0.4mm]
\multirow{6}{*}{\begin{turn}{90}Predicted Label\end{turn}} & \cellcolor{black!10}\textbf{\begin{tabular}[c]{@{}c@{}}1-label\\ Test$\rightarrow$\\ Train\\$ \downarrow$\end{tabular}} & \cellcolor{black!10}\textbf{CAVE} & \cellcolor{black!10}\textbf{MPII} & \cellcolor{black!10}\textbf{Gaze360} & \cellcolor{black!10}\textbf{TabletGaze}\\ \cline{2-6} \noalign{\vskip\doublerulesep \vskip-\arrayrulewidth} \cline{2-6}
                                          &      \textbf{CAVE}                                                           & --            & 0.55       &  0.54        & 0.50                    \\ 
                                          &      \textbf{MPII}                                                           & 0.30        & --            &   0.52       & 0.50                  \\ 
                                          &    \textbf{Gaze360}                                                        &  0.23     &  0.37      & --               &            0.49    \\  
                                          &  \textbf{TabletGaze}                                                     & 0.25        & 0.46         & 0.53          & --                           \\ 
                                          &   \textbf{YouTube}                   & 0.30        & 0.52        &  0.49          & 0.45                                                  \\ 
\bottomrule[0.4mm]
\end{tabular}
\vspace{-5mm}
\end{table*}

\subsection{Ablation Studies.}\label{sec:ablation}

\noindent \textbf{Impact of Loss Function.} We progressively integrate different parts of our method. We assess the impact of each loss term mentioned in Eq. \ref{eq:two_loss} and Eq. \ref{eq:one_loss} by considering them one at a time during learning on the MPII dataset. The results are shown in Table \ref{tab:ablation_loss}. For `2-labels' and `1-label', if only regression loss is considered, the error is high for the two techniques (0.98 and 0.99). We argue that this could be due to lack of domain knowledge. Further, consistency loss is added to the network, which reduces the error (0.42 and 0.76) significantly for both settings. On the other hand, KL divergence based loss is added to the `1-label' framework, which again reduces the error significantly from 0.76 to 0.58. Additionally, embedding loss is introduced to consider future frame consistency, though we note that for MPII, the change in MAE is very less (i.e. from 0.58 to 0.57). These experiments clearly establish the individual importance of each of the proposed loss terms in our framework.

\begin{wraptable}{r}{0.5\linewidth}
\centering
\caption{\small Comparison of model based methods on YouTube data. MAE: Mean Absolute Error, AE: Angular Error.}
\label{tab:model}
\scalebox{0.9}{
\begin{tabular}{c||cc}
\toprule[0.4mm] 
\cellcolor{black!10} \textbf{Method} &\cellcolor{black!10} \textbf{MAE} & \cellcolor{black!10} \textbf{AE} \\
\hline 
2-labels & \textbf{0.24} & \textbf{9.41} \\
~\cite{valenti2011combining} & 0.74 & 15.30 \\
~\cite{yamazoe2008remote} & 0.52 & 13.90 \\
~\cite{ishikawa2004passive} & 0.57 & 14.01 \\
\bottomrule[0.4mm]
\end{tabular}}
\end{wraptable}
\noindent \textbf{Impact of Motion Feature.} To judge the impact of the motion feature, we evaluate the performance of `2-labels' on the CAVE dataset without using the motion feature i.e. the latent space features are directly concatenated (Refer Fig.~\ref{fig:pipeline}). The results suggests that without the motion feature the MAE is reduced from 0.34 to 0.25. Thus, it is important to include this information in the framework.

\noindent \textbf{Sequential Modelling Vs `2-labels'.} We incorporate an LSTM module having 3 steps instead of the whole pipeline. The input to the LSTM module is 3-frame set and it estimate the gaze of the unlabelled frame. The MAE is quite high (i.e. 0.95) as compared to our ResNet-50 based `2-labels' framework (i.e. 0.25). The possible reason for this performance enhancement owes to task-relevant losses posed within very small temporal information.

\noindent \textbf{`Resnet-50+FC' Vs `2-labels'.} We also evaluate our method against a simple `ResNet+FC' trained on training partition and tested on the test partition. The MAE is 0.39 as compared to our ResNet-50 based `2-labels' framework (i.e. 0.25). This experiment also indicates the advantage of using triplet module.

\begin{figure}[t]
    \centering
    \includegraphics[width=\linewidth]{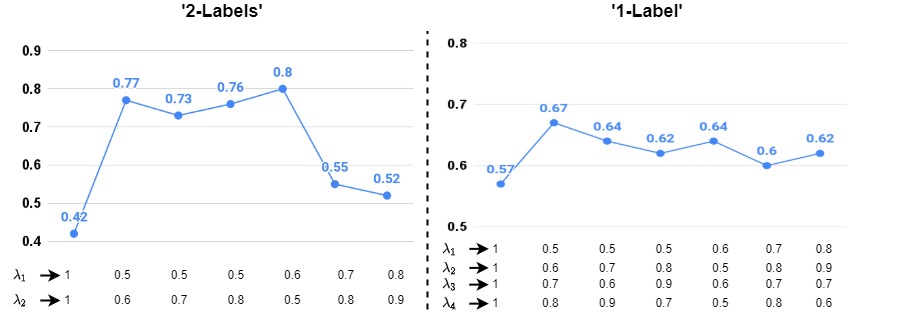}
    \caption{Impact of regularization parameters in `2-labels' and `1-label' settings.}
    \label{fig:regularizer}
    \vspace{-5mm}
\end{figure}

\noindent \textbf{Regularization Parameters.} We have also experimented with different values of the regularization parameters. The trade-off is shown in Fig.~\ref{fig:regularizer}. In this figure, the overall loss is plotted against different values of $\lambda$. It indicates that the optimal setting is achieved when all values are 1.

\subsection{Failure Cases} 
\begin{figure}[b]
\centering
\includegraphics[height = 0.75 in]{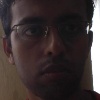}
\includegraphics[height = 0.75 in]{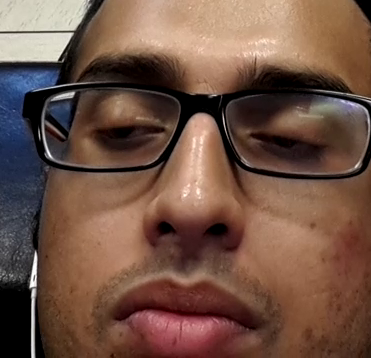}
\includegraphics[height = 0.75 in]{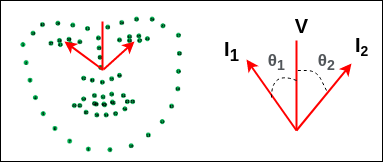}
\caption{  (Left) Two sample images for which our methods generate noisy labels due to illumination conditions and eye openness. (Right) Heuristic for gaze trajectory selection.}  
\label{fig:failure}
\vspace{-5mm}
\end{figure}

We also investigate the failure cases of our methods. The generated labels are noisy when the illumination is dark and the eyes are not open. Fig. \ref{fig:failure} shows a few cases, where the correlation is low as compared to the ground truth labels.

\noindent \textbf{Gaze Labelling `in the wild'.}  
The best results on the collected data in terms of MAE and CC are mentioned in the last row of Table \ref{tab:results}. The angular errors w.r.t. the SLERP~\cite{peters2010head} for `2-labels' and `1-label' are 9.41\textdegree and 12.07\textdegree, although SLERP is a weak baseline to compare eye gaze in an unconstrained environment. When we use the generated labels on YouTube dataset to complement the other dataset, the performance improves for TabletGaze and Gaze360 (see Table~\ref{tab:sota} and~\ref{tab:sota_1}). For adapting the SOTA methods (Table~\ref{tab:sota} and~\ref{tab:sota_1}), we fine-tune the models following standard protocol~\cite{dubey2019unsupervised,yu2019improving}. The results suggest that the network learns a meaningful representation. Moreover, we compare our method with model based approaches~\cite{valenti2011combining,yamazoe2008remote,ishikawa2004passive}. The results are depicted in Table~\ref{tab:model}. From the table, it is observed that our proposed method outperforms model based methods.

\begin{figure}[t]
    \centering
    \includegraphics[width = 0.8\linewidth]{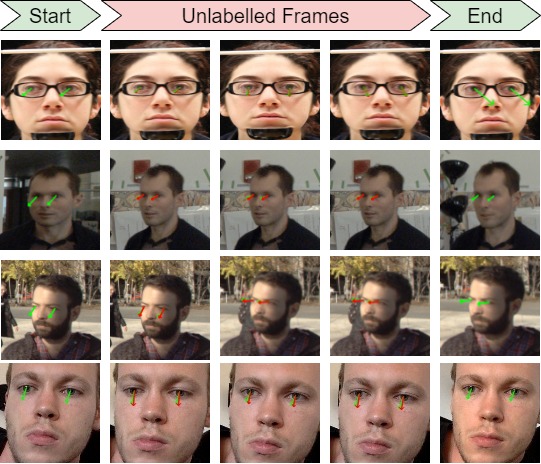}  
    \caption{\small Few examples of gaze trajectories along with qualitative prediction results. Here, red and green arrow represent predicted and ground truth gaze direction, respectively.}
    \label{fig:qualitative}
    \vspace{-5mm}
\end{figure}

\subsection{Qualitative Analysis.}\label{sec:qualitative}
Fig.~\ref{fig:qualitative} illustrates few examples of gaze trajectories along with the predictions of the proposed method. The trajectories consist of start, unlabelled and end frames and the trajectory length is not limited to 3. During training, the proposed method takes start, end and \textit{one of the unlabelled frame} as input. Please note that the eye patch cropped from the facial images are used as input. In the `2-labels' case, the ground truth labels of both start and end frames are provided for weak supervision. In Fig.~\ref{fig:qualitative}, the green and red arrow indicates ground truth and predicted gaze direction. For better understanding, we plot the gaze direction corresponding to each eye originated from detected pupil center. This qualitative analysis indicates that our weakly supervised method learns to interpolate gaze labels efficiently from terminal frames of a trajectory.

\section{Limitations, Conclusion and Future Work}
\label{sec:conclusion}
The main limitation of our study is the fixed gaze trajectory requirement. Moreover, we did not remove eye blink frames which leads to most of the failure cases. 
Another limitation of our framework is that it only considers mostly frontal faces (in terms of yaw and pitch axis of headpose). Our study introduces a weakly-supervised approach for generating labels for intermediate frames in a defined gaze trajectory. The proposed frameworks leverage task-specific domain knowledge i.e. trajectory ordering, motion and appearance features etc. With extensive experiments, we show that the labels generated by our methods are comparable to the ground truth labels. Further, we also show that the state-of-the-art existing techniques re-trained using the labels generated by our method give comparable performance. This applies that with just 1\%-5.6\% labelled data (dependent on the dataset) training can be performed with performance comparable to when 100\% training data is available. Further, we also propose a technique to collect and label eye gaze `in the wild'. The proposed method can be used for other computer vision based applications (e.g. gaze tracking devices for AR and VR) without the prior need of having to use the whole labelled dataset during training. In the future, we will investigate subject specific gaze estimation in challenging situations such as low-resolution images, uneven illumination conditions and extreme head-poses. Moreover, it will be interesting to explore extrapolation from the given gaze labels. Although our methods consider gaze annotation in few of the aforementioned diverse conditions, it would be interesting to have more in-depth study in this domain.

\bibliographystyle{splncs}
\bibliography{arxiv}

\begin{thebibliography}{10}

\bibitem{liu2011visual}
Liu, H., Heynderickx, I.:
\newblock Visual attention in objective image quality assessment: Based on
  eye-tracking data.
\newblock IEEE Transactions on Circuits and Systems for Video Technology
  \textbf{21} (2011)  971--982

\bibitem{wang2017deep}
Wang, W., Shen, J.:
\newblock Deep visual attention prediction.
\newblock IEEE Transactions on Image Processing \textbf{27} (2017)  2368--2378

\bibitem{mustafa2018prediction}
Mustafa, A., Kaur, A., Mehta, L., Dhall, A.:
\newblock Prediction and localization of student engagement in the wild.
\newblock arXiv preprint arXiv:1804.00858 (2018)

\bibitem{ghosh2020speak2label}
Ghosh, S., Dhall, A., Sharma, G., Gupta, S., Sebe, N.:
\newblock Speak2label: Using domain knowledge for creating a large scale driver
  gaze zone estimation dataset.
\newblock arXiv preprint arXiv:2004.05973 (2020)

\bibitem{gaze360_2019}
Kellnhofer, P., Recasens, A., Stent, S., Matusik, W., Torralba, A.:
\newblock Gaze360: Physically unconstrained gaze estimation in the wild.
\newblock In: IEEE International Conference on Computer Vision. (2019)

\bibitem{ghosh2021Automatic}
Ghosh, S., Dhall, A., Hayat, M., Knibbe, J., Ji, Q.:
\newblock Automatic gaze analysis: A survey of deep learning based approaches.
\newblock arXiv preprint arXiv:2108.05479 (2021)

\bibitem{niehorster2020glassesviewer}
Niehorster, D.C., Hessels, R.S., Benjamins, J.S.:
\newblock Glassesviewer: Open-source software for viewing and analyzing data
  from the tobii pro glasses 2 eye tracker.
\newblock Behavior Research Methods (2020)  1--10

\bibitem{zhang2017everyday}
Zhang, X., Sugano, Y., Bulling, A.:
\newblock Everyday eye contact detection using unsupervised gaze target
  discovery.
\newblock In: ACM User Interface Software and Technology. (2017)  193--203

\bibitem{smith2013gaze}
Smith, B., Yin, Q., Feiner, S., Nayar, S.:
\newblock Gaze locking: passive eye contact detection for human-object
  interaction.
\newblock In: ACM User Interface Software \& Technology. (2013)

\bibitem{kothari2021weakly}
Kothari, R., De~Mello, S., Iqbal, U., Byeon, W., Park, S., Kautz, J.:
\newblock Weakly-supervised physically unconstrained gaze estimation.
\newblock In: Proceedings of the IEEE/CVF Conference on Computer Vision and
  Pattern Recognition. (2021)  9980--9989

\bibitem{ghosh2022mtgls}
Ghosh, S., Hayat, M., Dhall, A., Knibbe, J.:
\newblock Mtgls: Multi-task gaze estimation with limited supervision.
\newblock In: Proceedings of the IEEE/CVF Winter Conference on Applications of
  Computer Vision. (2022)  3223--3234

\bibitem{yu2019unsupervised}
Yu, Y., Odobez, J.:
\newblock Unsupervised representation learning for gaze estimation.
\newblock IEEE Conference on Computer Vision and Pattern Recognition (2020)
  1--13

\bibitem{dubey2019unsupervised}
Dubey, N., Ghosh, S., Dhall, A.:
\newblock Unsupervised learning of eye gaze representation from the web.
\newblock In: 2019 International Joint Conference on Neural Networks (IJCNN),
  IEEE (2019)  1--7

\bibitem{swaminathan2018enabling}
Swaminathan, A., Ramachandran, M.:
\newblock Enabling augmented reality using eye gaze tracking (2018) US Patent
  9,996,150.

\bibitem{blattgerste2018advantages}
Blattgerste, J., Renner, P., Pfeiffer, T.:
\newblock Advantages of eye-gaze over head-gaze-based selection in virtual and
  augmented reality under varying field of views.
\newblock In: Proceedings of the Workshop on Communication by Gaze Interaction.
  (2018)  1--9

\bibitem{gumilar2021connecting}
Gumilar, I., Barde, A., Hayati, A.F., Billinghurst, M., Lee, G., Momin, A.,
  Averill, C., Dey, A.:
\newblock Connecting the brains via virtual eyes: Eye-gaze directions and
  inter-brain synchrony in vr.
\newblock In: Extended Abstracts of the 2021 CHI Conference on Human Factors in
  Computing Systems. (2021)  1--7

\bibitem{park2021talking}
Park, W., Heo, J., Lee, J.:
\newblock Talking through the eyes: User experience design for eye gaze
  redirection in live video conferencing.
\newblock In: International Conference on Human-Computer Interaction, Springer
  (2021)  75--88

\bibitem{Alonso_Mart_n_2012}
Alonso-Martín, F., Gorostiza, J.F., Malfaz, M., Salichs, M.A.:
\newblock User localization during human-robot interaction.
\newblock Sensors \textbf{12} (2012)  9913–9935

\bibitem{zabala2021modeling}
Zabala, U., Rodriguez, I., Mart{\'\i}nez-Otzeta, J.M., Lazkano, E.:
\newblock Modeling and evaluating beat gestures for social robots.
\newblock Multimedia Tools and Applications (2021)  1--18

\bibitem{park2018deep}
Park, S., Spurr, A., Hilliges, O.:
\newblock Deep pictorial gaze estimation.
\newblock In: European Conference on Computer Vision. (2018)  721--738

\bibitem{lu2017appearance}
Lu, F., Chen, X., Sato, Y.:
\newblock Appearance-based gaze estimation via uncalibrated gaze pattern
  recovery.
\newblock IEEE Transactions on Image Processing \textbf{26} (2017)  1543--1553

\bibitem{zhang2015appearance}
Zhang, X., Sugano, Y., Fritz, M., Bulling, A.:
\newblock Appearance-based gaze estimation in the wild.
\newblock In: IEEE Computer Vision and Pattern Recognition. (2015)  4511--4520

\bibitem{lu2015gaze}
Lu, F., Sugano, Y., Okabe, T., Sato, Y.:
\newblock Gaze estimation from eye appearance: A head pose-free method via eye
  image synthesis.
\newblock IEEE Transactions on Image Processing \textbf{24} (2015)  3680--3693

\bibitem{krafka2016eye}
Krafka, K., Khosla, A., Kellnhofer, P., Kannan, H., Bhandarkar, S., Matusik,
  W., Torralba, A.:
\newblock Eye tracking for everyone.
\newblock In: IEEE Computer Vision and Pattern Recognition. (2016)  2176--2184

\bibitem{zhang2017s}
Zhang, X., Sugano, Y., Fritz, M., Bulling, A.:
\newblock It’s written all over your face: Full-face appearance-based gaze
  estimation.
\newblock In: IEEE Computer Vision and Pattern Recognition Workshop. (2017)

\bibitem{zhang2017mpiigaze}
Zhang, X., Sugano, Y., Fritz, M., Bulling, A.:
\newblock Mpiigaze: Real-world dataset and deep appearance-based gaze
  estimation.
\newblock IEEE Transactions on Pattern Analysis and Machine Intelligence (2017)

\bibitem{jyoti2018automatic}
Jyoti, S., Dhall, A.:
\newblock Automatic eye gaze estimation using geometric \& texture-based
  networks.
\newblock In: International Conference on Pattern Recognition, IEEE (2018)
  2474--2479

\bibitem{FischerECCV2018}
Fischer, T., Chang, H.J., Demiris, Y.:
\newblock {RT-GENE: Real-Time Eye Gaze Estimation in Natural Environments}.
\newblock In: European Conference on Computer Vision. (2018)  339--357

\bibitem{sugano2014learning}
Sugano, Y., Matsushita, Y., Sato, Y.:
\newblock Learning-by-synthesis for appearance-based 3d gaze estimation.
\newblock In: IEEE Computer Vision and Pattern Recognition. (2014)  1821--1828

\bibitem{wang2018hierarchical}
Wang, K., Zhao, R., Ji, Q.:
\newblock A hierarchical generative model for eye image synthesis and eye gaze
  estimation.
\newblock In: Proceedings of the IEEE Conference on Computer Vision and Pattern
  Recognition. (2018)  440--448

\bibitem{benfold2011unsupervised}
Benfold, B., Reid, I.:
\newblock Unsupervised learning of a scene-specific coarse gaze estimator.
\newblock In: IEEE International Conference on Computer Vision. (2011)
  2344--2351

\bibitem{park2019few}
Park, S., Mello, S.D., Molchanov, P., Iqbal, U., Hilliges, O., Kautz, J.:
\newblock Few-shot adaptive gaze estimation.
\newblock In: IEEE International Conference on Computer Vision. (2019)
  9368--9377

\bibitem{yu2019improving}
Yu, Y., Liu, G., Odobez, J.:
\newblock Improving few-shot user-specific gaze adaptation via gaze redirection
  synthesis.
\newblock In: IEEE Conference on Computer Vision and Pattern Recognition.
  (2019)  11937--11946

\bibitem{duchowski2017eye}
Duchowski, A.T., Duchowski, A.T.:
\newblock Eye tracking methodology: Theory and practice.
\newblock Springer (2017)

\bibitem{komogortsev2013automated}
Komogortsev, O.V., Karpov, A.:
\newblock Automated classification and scoring of smooth pursuit eye movements
  in the presence of fixations and saccades.
\newblock Behavior research methods \textbf{45} (2013)  203--215

\bibitem{startsev20191d}
Startsev, M., Agtzidis, I., Dorr, M.:
\newblock 1d cnn with blstm for automated classification of fixations,
  saccades, and smooth pursuits.
\newblock Behavior Research Methods \textbf{51} (2019)  556--572

\bibitem{santini2016bayesian}
Santini, T., Fuhl, W., K{\"u}bler, T., Kasneci, E.:
\newblock Bayesian identification of fixations, saccades, and smooth pursuits.
\newblock In: Proceedings of the Ninth Biennial ACM Symposium on Eye Tracking
  Research \& Applications. (2016)  163--170

\bibitem{zhu2020hierarchical}
Zhu, Y., Yan, Y., Komogortsev, O.:
\newblock Hierarchical hmm for eye movement classification.
\newblock In: European Conference on Computer Vision, Springer (2020)  544--554

\bibitem{arabadzhiyska2017saccade}
Arabadzhiyska, E., Tursun, O.T., Myszkowski, K., Seidel, H.P., Didyk, P.:
\newblock Saccade landing position prediction for gaze-contingent rendering.
\newblock ACM Transactions on Graphics (TOG) \textbf{36} (2017)  1--12

\bibitem{huang2015tabletgaze}
Huang, Q., Veeraraghavan, A., Sabharwal, A.:
\newblock Tabletgaze: dataset and analysis for unconstrained appearance-based
  gaze estimation in mobile tablets.
\newblock Machine Vision and Applications \textbf{28} (2017)  445--461

\bibitem{zhang2020eth}
Zhang, X., Park, S., Beeler, T., Bradley, D., Tang, S., Hilliges, O.:
\newblock Eth-xgaze: A large scale dataset for gaze estimation under extreme
  head pose and gaze variation.
\newblock In: European Conference on Computer Vision, Springer (2020)  365--381

\bibitem{FunesMora_ETRA_2014}
Funes~Mora, K.A., Monay, F., Odobez, J.M.:
\newblock Eyediap: A database for the development and evaluation of gaze
  estimation algorithms from rgb and rgb-d cameras.
\newblock In: ACM Symposium on Eye Tracking Research and Applications. (2014)

\bibitem{garbin2019openeds}
Garbin, S.J., Shen, Y., Schuetz, I., Cavin, R., Hughes, G., Talathi, S.S.:
\newblock Openeds: Open eye dataset.
\newblock arXiv preprint arXiv:1905.03702 (2019)

\bibitem{palmero2020openeds2020}
Palmero, C., Sharma, A., Behrendt, K., Krishnakumar, K., Komogortsev, O.V.,
  Talathi, S.S.:
\newblock Openeds2020: Open eyes dataset.
\newblock arXiv preprint arXiv:2005.03876 (2020)

\bibitem{ganin2016deepwarp}
Ganin, Y., Kononenko, D., Sungatullina, D., Lempitsky, V.:
\newblock Deepwarp: Photorealistic image resynthesis for gaze manipulation.
\newblock In: European conference on computer vision, Springer (2016)  311--326

\bibitem{he2019photo}
He, Z., Spurr, A., Zhang, X., Hilliges, O.:
\newblock Photo-realistic monocular gaze redirection using generative
  adversarial networks.
\newblock In: Proceedings of the IEEE International Conference on Computer
  Vision. (2019)  6932--6941

\bibitem{wood2018gazedirector}
Wood, E., Baltru{\v{s}}aitis, T., Morency, L.P., Robinson, P., Bulling, A.:
\newblock Gazedirector: Fully articulated eye gaze redirection in video.
\newblock In: Computer Graphics Forum. Volume~37., Wiley Online Library (2018)
  217--225

\bibitem{kononenko2015learning}
Kononenko, D., Lempitsky, V.:
\newblock Learning to look up: Realtime monocular gaze correction using machine
  learning.
\newblock In: Proceedings of the IEEE Conference on Computer Vision and Pattern
  Recognition. (2015)  4667--4675

\bibitem{sela2017gazegan}
Sela, M., Xu, P., He, J., Navalpakkam, V., Lagun, D.:
\newblock Gazegan-unpaired adversarial image generation for gaze estimation.
\newblock arXiv preprint arXiv:1711.09767 (2017)

\bibitem{bilen2016weakly}
Bilen, H., Vedaldi, A.:
\newblock Weakly supervised deep detection networks.
\newblock In: IEEE Computer Vision and Pattern Recognition. (2016)  2846--2854

\bibitem{lee2013pseudo}
Lee, D.H.:
\newblock Pseudo-label: The simple and efficient semi-supervised learning
  method for deep neural networks.
\newblock In: International Conference on Machine Learning Workshop. Volume~3.
  (2013) ~2

\bibitem{zhang2018weakly}
Zhang, Y., Dong, W., Hu, B.G., Ji, Q.:
\newblock Weakly-supervised deep convolutional neural network learning for
  facial action unit intensity estimation.
\newblock In: IEEE Computer Vision and Pattern Recognition. (2018)  2314--2323

\bibitem{arandjelovic2016netvlad}
Arandjelovic, R., Gronat, P., Torii, A., Pajdla, T., Sivic, J.:
\newblock Netvlad: Cnn architecture for weakly supervised place recognition.
\newblock In: IEEE Computer Vision and Pattern Recognition. (2016)  5297--5307

\bibitem{zhang2018bilateral}
Zhang, Y., Zhao, R., Dong, W., Hu, B.G., Ji, Q.:
\newblock Bilateral ordinal relevance multi-instance regression for facial
  action unit intensity estimation.
\newblock In: Proceedings of the IEEE Conference on Computer Vision and Pattern
  Recognition. (2018)  7034--7043

\bibitem{weston2012deep}
Weston, J., Ratle, F., Mobahi, H., Collobert, R.:
\newblock Deep learning via semi-supervised embedding.
\newblock In: Neural networks: Tricks of the trade.
\newblock Springer (2012)  639--655

\bibitem{zhao2015stacked}
Zhao, J., Mathieu, M., Goroshin, R., Lecun, Y.:
\newblock Stacked what-where auto-encoders.
\newblock International Conference on Learning Representations Workshop (2015)

\bibitem{sajjadi2016mutual}
Sajjadi, M., Javanmardi, M., Tasdizen, T.:
\newblock Mutual exclusivity loss for semi-supervised deep learning.
\newblock In: IEEE International Conference on Image Processing. (2016)
  1908--1912

\bibitem{williams2006sparse}
Williams, O., Blake, A., Cipolla, R.:
\newblock Sparse and semi-supervised visual mapping with the s3gp.
\newblock In: IEEE Computer Vision and Pattern Recognition. (2006)

\bibitem{haeusser2017learning}
Haeusser, P., Mordvintsev, A., Cremers, D.:
\newblock Learning by association--a versatile semi-supervised training method
  for neural networks.
\newblock In: IEEE Computer Vision and Pattern Recognition. (2017)  89--98

\bibitem{purves2015perception}
Purves, D., Morgenstern, Y., Wojtach, W.T.:
\newblock Perception and reality: why a wholly empirical paradigm is needed to
  understand vision.
\newblock Frontiers in systems neuroscience \textbf{9} (2015)  156

\bibitem{majaranta2011gaze}
Majaranta, P.:
\newblock Gaze Interaction and Applications of Eye Tracking: Advances in
  Assistive Technologies: Advances in Assistive Technologies.
\newblock IGI Global (2011)

\bibitem{simonyan2014very}
Simonyan, K., Zisserman, A.:
\newblock Very deep convolutional networks for large-scale image recognition.
\newblock International Conference on Learning Representations (2014)

\bibitem{he2016deep}
He, K., Zhang, X., Ren, S., Sun, J.:
\newblock Deep residual learning for image recognition.
\newblock In: IEEE Computer Vision and Pattern Recognition. (2016)  770--778

\bibitem{schmidt2020depth}
Schmidt, S., Bruder, G., Steinicke, F.:
\newblock Depth perception and manipulation in projection-based spatial
  augmented reality.
\newblock PRESENCE: Virtual and Augmented Reality \textbf{27} (2020)  242--256

\bibitem{bertasius2019learning}
Bertasius, G., Feichtenhofer, C., Tran, D., Shi, J., Torresani, L.:
\newblock Learning temporal pose estimation from sparsely-labeled videos.
\newblock arXiv preprint arXiv:1906.04016 (2019)

\bibitem{barz2020deep}
Barz, B., Denzler, J.:
\newblock Deep learning on small datasets without pre-training using cosine
  loss.
\newblock In: IEEE Winter Conference on Applications of Computer Vision. (2020)
   1371--1380

\bibitem{baltruvsaitis2016openface}
Baltru{\v{s}}aitis, T., Robinson, P., Morency, L.P.:
\newblock Openface: an open source facial behavior analysis toolkit.
\newblock In: IEEE Winter Conference on Applications of Computer Vision. (2016)
   1--10

\bibitem{eberly2011fast}
Eberly, D.:
\newblock A fast and accurate algorithm for computing slerp.
\newblock Journal of Graphics, GPU, and Game Tools \textbf{15} (2011)  161--176

\bibitem{valenti2011combining}
Valenti, R., Sebe, N., Gevers, T.:
\newblock Combining head pose and eye location information for gaze estimation.
\newblock IEEE Transactions on Image Processing \textbf{21} (2011)  802--815

\bibitem{yamazoe2008remote}
Yamazoe, H., Utsumi, A., Yonezawa, T., Abe, S.:
\newblock Remote gaze estimation with a single camera based on facial-feature
  tracking without special calibration actions.
\newblock In: Proceedings of the 2008 symposium on Eye tracking research \&
  applications. (2008)  245--250

\bibitem{ishikawa2004passive}
Ishikawa, T.:
\newblock Passive driver gaze tracking with active appearance models.
\newblock (2004)

\bibitem{park2018learning}
Park, S., Zhang, X., Bulling, A., Hilliges, O.:
\newblock Learning to find eye region landmarks for remote gaze estimation in
  unconstrained settings.
\newblock In: Proceedings of the 2018 ACM Symposium on Eye Tracking Research \&
  Applications. (2018)  1--10

\bibitem{sharma2016farec}
Sharma, S., Shanmugasundaram, K., Ramasamy, S.K.:
\newblock Farec—cnn based efficient face recognition technique using dlib.
\newblock In: International Conference on Advanced Communication Control and
  Computing Technologies. (2016)  192--195

\bibitem{github_gaze}
:
\newblock gaze code:\url{https://github.com/swook/GazeML,
  https://github.com/Erkil1452/gaze360} (-)

\bibitem{peters2010head}
Peters, C., Qureshi, A.:
\newblock A head movement propensity model for animating gaze shifts and blinks
  of virtual characters.
\newblock Computers \& Graphics \textbf{34} (2010)  677--687

\end{thebibliography}

\end{document}